%% file: main.tex
\documentclass[lettersize,journal]{IEEEtran}
\input{math_command.tex}

\usepackage{amsmath,amsfonts}
\usepackage{amssymb}
\usepackage{algorithmic}
\usepackage{algorithm}
\usepackage{array}
\usepackage[caption=false,font=normalsize,labelfont=sf,textfont=sf]{subfig}
\usepackage{textcomp}
\usepackage{stfloats}
\usepackage{url}
\usepackage{verbatim}
\usepackage{graphicx}

\usepackage{cite}
\usepackage{caption}
\usepackage{makecell}
\usepackage{contour}
\usepackage{wrapfig}
\usepackage{graphicx}%
\usepackage{multirow}%
\usepackage{booktabs}
\usepackage{adjustbox}
\usepackage{tcolorbox}
\PassOptionsToPackage{table}{xcolor}
\usepackage{booktabs}  %
\usepackage{adjustbox} %
\usepackage{colortbl}  %
\usepackage[table]{xcolor} %
\usepackage{hyperref}

\definecolor{mygray}{gray}{.95}
\definecolor{mypink}{rgb}{.99,.91,.95}
\definecolor{mycyan}{cmyk}{.3,0,0,0}
\definecolor{darkgreen}{rgb}{0.0, 0.5, 0.0}
\makeatletter
  \newcommand\figcaption{\def\@captype{figure}\caption}
  \newcommand\tabcaption{\def\@captype{table}\caption}
  
\makeatother
\definecolor{green(munsell)}{rgb}{0.0, 0.66, 0.47}
\definecolor{beaublue}{rgb}{0.74, 0.83, 0.9}
\definecolor{bluegray}{rgb}{0.4, 0.6, 0.8}
\definecolor{airforceblue}{rgb}{0.35, 0.5, 0.72}
\makeatletter
\renewcommand{\paragraph}{%
  \@startsection{paragraph}{4}%
  {\z@}{0.5em}{-0.0em}%
  {\normalfont\normalsize\bfseries}%
}
\makeatother

\hypersetup{
    colorlinks=true,   %
    citecolor=green(munsell),    %
    linkcolor=blue,   %
    urlcolor=green(munsell)      %
}

\begin{document}

\title{Confidence and Dispersity as Signals: Unsupervised Model Evaluation and Ranking}

\author{Weijian Deng, Weijie Tu, Ibrahim Radwan, Mohammad Abu Alsheikh, Stephen Gould, Liang Zheng
\thanks{W. Tu, W. Deng, S. Gould, and L. Zheng are with the School of Computing, The Australian National University, Canberra, ACT 0200, Australia. 
E-mail: \{firstname.lastname\}@anu.edu.au. %
I. Radwan and M. A. Alsheikh are with the University of Canberra.
E-mail: \{ibrahim.radwan, Mohammad.Abualsheikh\}@canberra.edu.au%
}}

\markboth{Journal of \LaTeX\ Class Files,~Vol.~14, No.~8, August~2015}%
{Shell \MakeLowercase{\textit{et al.}}: Bare Advanced Demo of IEEEtran.cls for IEEE Computer Society Journals}

\maketitle
  
\begin{abstract}
Assessing model generalization under distribution shift is essential for real-world deployment, particularly when labeled test data is unavailable. This paper presents a unified and practical framework for unsupervised model evaluation and ranking in two common deployment settings: (1) estimating the accuracy of a fixed model on multiple unlabeled test sets (dataset-centric evaluation), and (2) ranking a set of candidate models on a single unlabeled test set (model-centric evaluation). We demonstrate that two intrinsic properties of model predictions, namely confidence (which reflects prediction certainty) and dispersity (which captures the diversity of predicted classes), together provide strong and complementary signals for generalization. We systematically benchmark a set of confidence-based, dispersity-based, and hybrid metrics across a wide range of model architectures, datasets, and distribution shift types. Our results show that hybrid metrics consistently outperform single-aspect metrics on both dataset-centric and model-centric evaluation settings. In particular, the nuclear norm of the prediction matrix provides robust and accurate performance across tasks, including real-world datasets, and maintains reliability under moderate class imbalance. These findings offer a practical and generalizable basis for unsupervised model assessment in deployment scenarios.
\end{abstract}


\section{Introduction}
\IEEEPARstart{M}{odel} evaluation is essential for validating, selecting, and deploying machine learning systems~\cite{torralba2024foundations}. Conventionally, this is done on labeled validation or test sets drawn from the same distribution as the training data~\cite{arlot2010survey}. However, such an assumption rarely holds in real-world applications~\cite{djolonga2021robustness,koh2021wilds,kirsch2022note}, where models encounter data from dynamic and unknown environments.
In autonomous driving, for instance, models must operate under diverse conditions, nighttime, rain, and unusual traffic patterns, yet it is costly and time-consuming to label data from every possible setting. Even when labels are available, they often represent only a narrow slice of the real world, introducing evaluation bias.

This challenge, \textit{i.e.}, evaluating models without labeled data, has motivated growing interest in unsupervised model evaluation, which aims to estimate the accuracy of a trained model on an unlabeled test set~\cite{deng2021labels,guillory2021predicting,Deng:ICML2021,garg2022leveraging,baek2022agreement,yu2022predicting,chen2021mandoline,chen2021detecting}. Without access to ground-truth labels, existing methods rely on internal signals, especially the distribution of prediction confidences~\cite{hendrycks2016baseline, guillory2021predicting, garg2022leveraging, deng2023confidence}. Several works use summary statistics of model outputs, such as the average maximum softmax score~\cite{guillory2021predicting,hendrycks2016baseline} or prediction entropy~\cite{guillory2021predicting}, as proxies for generalization. These confidence-based metrics capture how certain a model is about its predictions on individual samples.

However, confidence alone does not always reflect true generalization. 
A model may be consistently confident yet predict only a small subset of classes, indicating limited adaptability under distribution shift. We study an additional perspective: prediction dispersity, which quantifies how predictions are distributed across all classes. A well-generalizing model should not only be confident in individual samples but also produce diverse predictions over the test set. Confidence characterizes sample-level certainty, while dispersity captures set-level diversity and class sensitivity.
To jointly capture both properties, our conference version uses the nuclear norm of the prediction matrix~\cite{deng2023confidence}. It aggregates the softmax outputs across the test set and summarizes both certainty and distributional spread. Empirically, the nuclear norm demonstrates robust performance across various benchmarks, outperforming confidence-only methods under distribution shift.

Building upon this insight, we extend our investigation to a complementary generalization analysis task: unsupervised model ranking. Rather than evaluating a single model across datasets, the goal here is to rank a pool of candidate models by their expected performance on a given, unlabeled test set. This setting frequently arises in practice, such as when choosing between architectures, training variants, or fine-tuned models for deployment in a new domain.

\begin{figure*}[!ht]
    \centering
\includegraphics[width=1\linewidth]{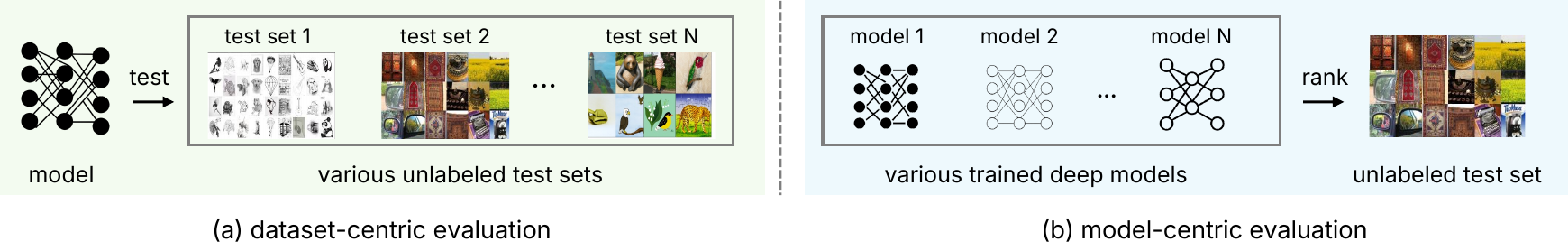}
    \caption{
    \textbf{Illustration of the two unsupervised generalization analysis tasks.}
(a) Dataset-Centric Evaluation: A fixed model is evaluated on a collection of unlabeled test sets drawn from diverse distributions. The objective is to estimate its generalization performance on each distribution without access to labeled data.
(b) Model-Centric Evaluation: A single unlabeled test set is used to compare multiple candidate models. The goal is to rank the models by their expected performance on the target distribution without relying on test-time supervision. These two complementary setups enable a comprehensive understanding of model generalization under distribution shift in a label-free manner.}
\label{fig:framework}
\end{figure*}

We refer to the two settings illustrated in Fig.~\ref{fig:framework} as follows:
(1) Dataset-centric evaluation, where the objective is to estimate the accuracy of a fixed model across multiple unlabeled test datasets that may differ in distribution;
(2) Model-centric evaluation, where the objective is to identify the most suitable model from a set of candidates for a single unlabeled test dataset.
Despite their structural differences, both tasks share the fundamental challenge of predicting model generalization performance in the absence of ground-truth labels.

In the experiments, we systematically investigate unsupervised metrics for assessing model generalization on unlabeled data. We consider three categories of metrics: confidence-based, dispersity-based, and hybrid metrics that capture both properties. We benchmark these metrics on both evaluation and ranking tasks across diverse datasets, architectures, and types of distribution shift.
Our key finding is that metrics that jointly consider confidence and dispersity provide more robust and reliable estimates of generalization. Models that produce predictions that are both confident on individual samples and well-distributed across classes tend to generalize better, both in dataset-centric evaluation and in comparative ranking.	
Moreover, NuclearNorm demonstrates robust performance across both dataset-centric evaluation and model-centric ranking tasks, consistently outperforming other metrics. While hybrid approaches may face limitations under severe class imbalance, our analysis reveals their resilience in moderately imbalanced settings.

To summarize, our contributions are as follows:
\begin{itemize}
    \item \textbf{Unified perspective on unsupervised generalization analysis.}  
    We study two key tasks—dataset-centric evaluation and model-centric ranking—that both aim to assess model generalization without labeled data, highlighting their differences and shared challenges.

    \item \textbf{Comprehensive evaluation of unsupervised metrics.}  
    We systematically evaluate a wide range of metrics across three categories: confidence-based, dispersity-based, and combined metrics. Our analysis spans diverse model architectures, datasets, and distribution shifts, revealing when different metrics are reliable.
    
    \item \textbf{Consistent behavior of combined metrics.}  
    Across the two key tasks, grouping metrics into confidence-based, dispersity-based, and hybrid-based shows that hybrid methods generally perform best, with nuclear norm the most robust and accurate.
\end{itemize}

This journal version substantially extends our conference paper~\cite{deng2023confidence} by introducing new evaluation settings, refined metric categorization, and expanded experimental analysis. First, we add a model-centric evaluation task (Sec.~\ref {sec:model}), where the goal is to rank multiple models on an unlabeled test set—an important yet underexplored setting for real-world deployment. Second, we propose a taxonomy that groups metrics into confidence-based, dispersity-based, and hybrid types, providing interpretability and clarifying the complementary nature of different signals (Sec.~\ref{sec:method}). Third, for the dataset-centric evaluation task (Sec.~\ref{sec:data}), we enhance the study by including recent methods (\textit{e.g.}, AvgEnergy~\cite{peng2024energybased}, MaNO~\cite{xie2024mano}, COT~\cite{cot2023}, SoftmaxCorr~\cite{tu2024what}), incorporating zero-shot vision-language models, and testing robustness under 3D-aware distribution shifts and class imbalance. Together, these contributions offer a more comprehensive and practically relevant analysis of unsupervised accuracy estimation.

\section{Related Work}

\noindent \textbf{Out-of-Distribution Generalization.}
A central goal in machine learning is to ensure that models trained on a source distribution perform reliably on unseen target distributions. 
Theoretical work has aimed to characterize and bound OOD generalization error. Foundational analyses~\cite{ben2006analysis, ben2010theory} derive upper bounds for domain adaptation settings. More recent efforts have connected generalization performance to measures of distributional divergence, including $f$-divergences and optimal transport distances~\cite{zhang2019bridging, acuna2021f}, offering deeper insights into model behavior across shifts.

\noindent \textbf{Predicting In-Distribution Generalization.} This task aims to estimate the generalization gap between training and test accuracy under the assumption that both sets share the same distribution~\cite{jiang2018predicting, neyshabur2017exploring, jiang2019fantastic, schiff2021predicting}. For example, the method in \cite{corneanu2020computing} computes topological descriptors to capture structural properties of learned characteristics, while \cite{jiang2018predicting} introduces a margin-based metric that analyzes the distribution of prediction margins across network layers.
While effective in controlled settings, these methods typically overlook test data properties and are not designed to handle real-world distribution shifts
In contrast, we address a more practical and challenging setting: unsupervised evaluation and ranking of models across diverse, out-of-distribution test sets.

\noindent \textbf{Unsupervised Accuracy Estimation.}
Estimating model accuracy on unlabeled test sets has emerged as a key problem for assessing generalization under distribution shift. Prior work explores several directions to tackle this challenge. One line of research leverages model outputs on test data, using summary statistics such as maximum confidence, entropy, or the shape of the softmax distribution to approximate accuracy~\cite{hendrycks2016baseline, guillory2021predicting, garg2022leveraging, deng2023confidence}. Our work addresses this issue by jointly modeling prediction confidence and class-wise coverage to improve robustness.
A second approach focuses on quantifying the distribution shift between training and test data~\cite{deng2021labels, yu2022predicting, cot2023}. They assume that larger shifts imply greater performance degradation, though the accuracy-discrepancy correlation is often inconsistent~\cite{guillory2021predicting, xie2023importance}, and some approaches incur high computational cost due to their reliance on training data~\cite{deng2021labels}.
Moreover, unsupervised loss-based techniques estimate accuracy using proxy signals such as self-supervised consistency~\cite{jiang2021assessing} or prediction agreement across multiple classifiers~\cite{madani2004co,platanios2016estimating,chen2021detecting}. However, many of these methods require access to multiple models or architectural changes, limiting their broad applicability. In contrast, our study focuses on standard softmax outputs from off-the-shelf classifiers, aiming to provide generalizable and training-free accuracy estimation by unifying confidence and dispersity signals.

\noindent \textbf{Confidence Calibration.}
Confidence calibration aims to align a model’s predicted confidence with the actual accuracy of samples at the same confidence level~\cite{guo2017calibration,minderer2021revisiting}. 
A well-calibrated model should have its average predicted confidence match the actual accuracy. However, many calibration approaches struggle to maintain this alignment under distribution shifts~\cite{ovadia2019can,gong2021confidence,zou2023adaptive}.
This work does not aim to calibrate confidence. Instead, we study how the confidence and dispersity of the prediction matrix can be used to analyze model generalization on unlabeled test sets.

\noindent \textbf{Out-of-Distribution Detection.} Out-of-distribution (OOD) detection focuses on identifying inputs from unseen classes that a model should ideally abstain from predicting~\cite{huang2021importance,du2024does}. Existing methods often rely on scoring functions derived from model outputs, including confidence-based~\cite{hendrycks2016baseline}, energy-based~\cite{liu2020energy}, logit-based~\cite{hendrycks2022scaling,wang2022vim}, and distance-based~\cite{sun2022dice,ming2024does} approaches.
While these methods also utilize model outputs, their objective is to detect and filter out inputs from unseen classes. In contrast, our goal is to assess model generalization on unlabeled OOD test sets.

\section{Characterizing Prediction Matrix for Generalization Analysis}\label{sec:method}
To evaluate and compare models without relying on test labels, we focus on analyzing their prediction matrices, which are the collections of output probabilities generated by a model on an unlabeled test set. We introduce key components and define three categories of metrics: confidence-based, dispersity-based, and hybrid.

\noindent \textbf{Prediction Matrix.}
Given a trained classifier $f: \mathbb{R}^d \to \mathbb{R}^k$ that maps an input to a $k$-dimensional logit vector, and an unlabeled target dataset $\mathcal{D}_\mathrm{test}^{T} = \{\mathbf{x}^t_i\}_{i=1}^{n_t}$ sampled i.i.d.\ from $p_T$, we define the \textit{prediction matrix} $\bm{P}_m \in \mathbb{R}^{n_t \times k}$ as the collection of softmax outputs on all target samples:
\[
\bm{P}_m = 
\begin{bmatrix}
\mathbf{p}^t_1 \\
\mathbf{p}^t_2 \\
\vdots \\
\mathbf{p}^t_{n_t}
\end{bmatrix}, \quad \text{where } \mathbf{p}^t_i = \sigma(f(\mathbf{x}^t_i)) \in \Delta_k.
\]
Here, $\sigma(\cdot)$ denotes the Softmax function: for logits $\mathbf{z} \in \mathbb{R}^k$, the $j$-th entry of $\sigma(\mathbf{z})$ is defined as
\[
\sigma(\mathbf{z})[j] = \frac{e^{\mathbf{z}[j]}}{\sum_{l=1}^{k} e^{\mathbf{z}[l]}}.
\]
All entries of $\bm{P}_m$ lie in $[0,1]$, and each row sums to one.

\vspace{1mm}
\noindent\textbf{I. Prediction Confidence} measures whether a softmax vector (each row of $\mP$) is certain. Common ways to quantify confidence include the maximum softmax score and the entropy of the distribution. If the overall confidence of $\mP$ is high, this implies that the classifier $f$ is confident in its predictions on the test set.  We introduce the metrics that measure the prediction confidence. All methods operate on the softmax outputs of a given classifier $f$ and the unlabeled test set $\mathcal{D}_\mathrm{test}^{T}$.

\vspace{1mm}
\noindent \underline{Average Confidence (ConfScore)~\cite{hendrycks2017baseline}.}  
It computes the mean of the maximum predicted probability across the test set:
\begin{equation}\label{eq:confscore}
\mathrm{ConfScore} = \frac{1}{n_t} \sum_{i=1}^{n_t} \max_{j \in \mathcal{Y}} \bm{p}^t_i[j],
\end{equation}
which reflects the average prediction confidence over the most probable class.

\vspace{1mm}
\noindent \underline{\textit{Average Negative Entropy (Entropy)}~\cite{guillory2021predicting}.} 
This metric captures the average uncertainty of the model's predictions. Lower values indicate more confident predictions:
\begin{equation}\label{eq:entropy}
\mathrm{Entropy} = -\frac{1}{n_t} \sum_{i=1}^{n_t} H(\bm{p}^t_i),
\end{equation}
where $H(\bm{p}) = -\sum_{j=1}^k \bm{p}[j] \log \bm{p}[j]$ is the Shannon entropy.

\vspace{1mm}
\noindent \underline{\textit{Average Thresholded Confidence (ATC)}~\cite{garg2022leveraging}.}  
A threshold $t$ is calibrated on a labeled source validation set $\mathcal{D}_\mathrm{val}^{S}$ such that:
\begin{equation}\label{eq:atc}
\frac{1}{n_v} \sum_{i=1}^{n_v} \mathbb{I} \left[ \max_j \bm{p}^v_i[j] > t \right] = \text{Acc}(f; \mathcal{D}_\mathrm{val}^{S}).
\end{equation}
The ATC score estimates target accuracy by computing the proportion of confident predictions above $t$:
\begin{equation}\label{eq:atc}
\mathrm{ATC} = \frac{1}{n_t} \sum_{i=1}^{n_t} \mathbb{I} \left[ \max_j \bm{p}^t_i[j] > t \right].
\end{equation}

\vspace{1mm}
\noindent \underline{\textit{Average Energy (AvgEnergy)}~\cite{liu2020energy,peng2024energybased}.}  
This method computes the average energy score from unnormalized logits $\bm{z}_i = f(\bm{x}^t_i)$:
\begin{equation}\label{eq:avgenergy}
\mathrm{AvgEnergy} = -\frac{1}{n_t} \sum_{i=1}^{n_t} T \cdot \log \left( \sum_{j=1}^{k} \exp \left( \frac{z_{ij}}{T} \right) \right),
\end{equation}
where $T$ is a temperature parameter. Higher energy values generally indicate lower prediction confidence.

\vspace{1mm}
\noindent \underline{\textit{Difference of Confidence (DoC)}~\cite{guillory2021predicting}.}  
Estimates performance by correcting the source validation accuracy using the confidence shift:
\begin{equation}\label{eq:doc}
\mathrm{DoC} = \text{Acc}(f; \mathcal{D}_\mathrm{val}^{S}) - \left( \mathrm{ConfScore}_{\mathcal{D}_\mathrm{val}^{S}} - \mathrm{ConfScore}_{\mathcal{D}_\mathrm{test}^{T}} \right).
\end{equation}

\vspace{1mm}
\noindent \underline{\textit{MaNo}~\cite{xie2024mano}.}  
It defines a piecewise normalization function over logits $\bm{z}_i = f(\mathbf{x}^t_i)$:
\[
v(\bm{z}_i) = 
\begin{cases}
1 + \bm{z}_i + \frac{1}{2} \bm{z}_i^2, & \text{if } \tau \leq \eta \\
\exp(\bm{z}_i), & \text{otherwise}
\end{cases}, \quad
\bm{q}_i = \frac{v(\bm{z}_i)}{\sum_j v(\bm{z}_i)_j} \in \Delta_k.
\]
Here $\tau$ is computed as the average KL divergence between the softmax predictions and the uniform distribution. With fixed $\eta=5$, and collecting $\bm{q}_i$ into $\bm{Q} \in \mathbb{R}^{n_t \times k}$, the final score is:
\begin{equation}\label{eq:lpnorm}
\mathrm{LpNormScore} = \left( \frac{1}{n_t k} \sum_{i=1}^{n_t} \sum_{j=1}^k |\bm{Q}_{ij}|^4 \right)^{\frac{1}{4}} \in [0,1].
\end{equation}

\noindent\textbf{II. Prediction Dispersity} assesses how evenly predictions are distributed across the $k$ classes. High dispersity suggests balanced class assignments, while low dispersity indicates concentration on a few classes. This often occurs under distribution shift, where target features form dominant clusters misaligned with the source domain~\cite{liang2020we,yang2021exploiting,tang2020unsupervised}. As a result, the model may overpredict certain classes and ignore others. We examine whether dispersity can serve as a useful signal for unsupervised model evaluation and ranking, and study two metrics to capture this property.

\vspace{1mm}
\noindent \underline{\textit{ClassEntropy}.}
It computes the entropy of the marginal (average) predicted distribution:
\begin{equation}\label{eq:ClassEntropy}
\mathrm{Dispersity} = H\left( \frac{1}{n_t} \sum_{i=1}^{n_t} \bm{p}^t_i \right).
\end{equation}

\vspace{1mm}
\noindent \underline{\textit{Class Transport Distance (CTD)}.}  
It compares the predicted target label distribution to a reference source distribution (\textit{i.e.}, the uniform distribution) using Wasserstein distance. Each prediction is converted to a one-hot vector based on its top class, forming an empirical histogram $\bm{h}_T$. Let $\bm{h}_S$ be the source label histogram. The CTD score is defined as:
\begin{equation}\label{eq:ctd}
\mathrm{CTD} = \min_{\bm{T} \in \Pi(\bm{h}_T, \bm{h}_S)} \sum_{i,j} \bm{T}_{ij} \cdot \|i - j\|_{\infty},
\end{equation}
where $\Pi(\bm{h}_T, \bm{h}_S)$ denotes transport plans with $\bm{h}_T$ and $\bm{h}_S$.

\noindent\textbf{III. Prediction Confidence and Dispersity.} Our key insight is that a well-performing model should yield predictions with high confidence and high dispersity. That is, we need to consider both properties so as to make more accurate estimates. We study the following metrics:

\vspace{1mm}
\noindent \underline{Information Maximization (IM)~\cite{liang2020we,yang2021exploiting,tang2020unsupervised}.}  It is computed as the difference between the entropy of the marginal distribution and the average entropy of individual predictions:
\begin{equation}\label{eq:im}
\mathrm{IM} = H\left( \frac{1}{n_t} \sum_{i=1}^{n_t} \bm{p}^t_i \right) - \frac{1}{n_t} \sum_{i=1}^{n_t} H(\bm{p}^t_i),
\end{equation}
where the Shannon entropy is defined as \( H(\bm{p}) = -\sum_{j=1}^{k} \bm{p}[j] \log \bm{p}[j] \). 
The first term reflects class-level dispersity, while the second term measures prediction uncertainty.

\vspace{1mm}
\noindent \underline{Nuclear Norm (NuclearNorm)~\cite{deng2023confidence}.}  
Given the prediction matrix $\bm{P}_m = [\bm{p}^t_1; \dots; \bm{p}^t_{n_t}] \in \mathbb{R}^{n_t \times k}$, this method measures the sum of singular values:
\begin{equation}\label{eq:nuclearnorm}
\mathrm{NuclearNorm} = \frac{\|\bm{P}_m\|_*}{\sqrt{\min(n_t, k) \cdot n_t}},
\end{equation}
which jointly captures the prediction confidence and the diversity (dispersity) of outputs across the dataset.

\vspace{1mm}
\noindent \underline{Confidence Optimal Transport (COT)~\cite{cot2023}.}  
This method models the prediction distribution over classes as a probability measure and computes its distance to a reference distribution (\textit{i.e.}, the uniform distribution) via Wasserstein distance:
\begin{equation}\label{eq:classcorr}
\mathrm{COT} = W_{\infty} \left( f_{\#} \mathcal{P}_T, \mathcal{P}_S \right),
\end{equation}
where $f_{\#} \mathcal{P}_T$ is the pushforward distribution of the target predictions and $W_\infty$ uses $\ell_\infty$ cost.

\vspace{1mm}
\noindent \underline{{SoftmaxCorr}~\cite{tu2024what}.}  
This metric evaluates how well the class-class correlation structure from model predictions aligns with a prior class distribution. The class correlation matrix is computed from the prediction matrix $\bm{P}_m \in \mathbb{R}^{n_t \times K}$ as:
\[
\bm{C} = \frac{1}{n_t} \bm{P}_m^\top \bm{P}_m,
\]
where $n_t$ is the number of test samples and $K$ is the number of classes. Given a reference diagonal matrix $\bm{R} = \mathrm{diag}(\bm{d})$, where $\bm{d}$ denotes a prior class distribution, the SoftmaxCorr score is the cosine similarity between $\bm{C}$ and $\bm{R}$:
\begin{equation}\label{eq:softmaxcorr}
\mathrm{SoftmaxCorr} = \frac{\langle \bm{C}, \bm{R} \rangle}{\|\bm{C}\|_F \cdot \|\bm{R}\|_F}.
\end{equation}
Following~\cite{tu2024what}, the prior distribution $\bm{d}$ is obtained by averaging zero-shot prediction probabilities over the test set using a vision-language model (ViT-bigG/14-CLIPA).

\begin{table}[t]
\centering
\small
\caption{Summary of prediction-based metrics used for unsupervised generalization analysis. Metrics are grouped by the properties they capture: confidence, dispersity, or both. We also indicate the expected correlation with accuracy: $\uparrow$ means positive correlation, $\downarrow$ means negative correlation.}
\label{tab:metrics}
\setlength{\tabcolsep}{10pt}
\begin{tabular}{ll}
\toprule
\textbf{Category} & \textbf{Metric (Expected Corr.)} \\
\midrule
\multirow{6}{*}{Confidence} 
& ConfScore~($\uparrow$; Eq.~\ref{eq:confscore}) \\
& Entropy~($\uparrow$; Eq.~\ref{eq:entropy}) \\
& ATC~($\uparrow$; Eq.~\ref{eq:atc}) \\
& AvgEnergy~($\uparrow$; Eq.~\ref{eq:avgenergy}) \\
& DoC~($\uparrow$; Eq.~\ref{eq:doc}) \\
& MaNo~($\uparrow$; Eq.~\ref{eq:lpnorm}) \\
\midrule
\multirow{2}{*}{Dispersity}
& ClassEntropy~($\uparrow$; Eq.~\ref{eq:ClassEntropy}) \\
& CTD~($\downarrow$; Eq.~\ref{eq:ctd}) \\
\midrule
\multirow{4}{*}{Confidence + Dispersity}
& NuclearNorm~($\uparrow$; Eq.~\ref{eq:nuclearnorm}) \\
& COT~($\downarrow$; Eq.~\ref{eq:classcorr}) \\
& SoftmaxCorr~($\uparrow$; Eq.~\ref{eq:softmaxcorr}) \\
& IM~($\uparrow$; Eq.~\ref{eq:im}) \\
\bottomrule
\end{tabular}
\end{table}

Table~\ref{tab:metrics} summarizes the proposed prediction-based metrics, categorized by the property they aim to capture: confidence, dispersity, or a combination of both. Each metric's expected correlation direction with model accuracy (positive or negative) is also indicated to facilitate interpretability. These metrics will be evaluated for both unsupervised model evaluation and model ranking tasks in the following sections.

\begin{table*}[t]
	\begin{center}
	 \caption{\textbf{Comparison of 12 unsupervised metrics across CIFAR-10, CUB-200, ImageNet-C, and ImageNet-3D in dataset-centric accuracy estimation task.} 
  We report the coefficient of determination ($R^2$) between each metric and ground-truth model accuracy under the dataset-centric evaluation setting. Metrics are grouped into three categories: confidence-based, dispersity-based, and hybrid. Confidence-based metrics such as ATC and DoC perform well on CIFAR-10 and ImageNet-C but show reduced effectiveness on CUB-200 and ImageNet-3D for certain architectures. Dispersity-based metrics, particularly CTD and ClassEntropy, provide relatively high correlations across architectures. Hybrid metrics, including NuclearNorm, COT, and IM, generally achieve the highest performance across setups.
The best, second-best, and third-best metrics in each row are highlighted in \textcolor{red}{red}, \textcolor{darkgreen}{green}, and \textcolor{blue}{blue}, respectively.
     }
	\label{tab:corr_rho}
	\setlength{\tabcolsep}{2pt}
	\small
        \begin{tabular}{cc|cccccc|cc|cccc}
    \toprule
    \multirow{2}{*}{\textbf{Setup}}& \multirow{2}{*}{\textbf{Model} }
    & \multicolumn{6}{c|}{\textbf{Confidence}} 
    & \multicolumn{2}{c|}{\textbf{Dispersity}} 
    & \multicolumn{4}{c}{\textbf{Confidence + Dispersity}} \\
    \cmidrule(lr){3-8} \cmidrule(lr){9-10} \cmidrule(lr){11-14}
    & & ConfScore & Entropy & ATC & AvgEnergy & DoC & MaNo 
      & ClassEntropy & CTD 
      & NuclearNorm & COT & SoftmaxCorr & IM \\
\midrule
 \multirow{5}{*}[3pt]{\rotatebox[origin=c]{90}{CIFAR-10}}  
& ResNet-20 & 0.924 & 0.923 & 0.931 & 0.944  & 0.936 & 0.922 & 0.946 & 0.963 & 0.989 & 0.985 & 0.954 & 0.992 \\
& RepVGG-A0 & 0.817 & 0.815 & 0.836 & 0.753  & 0.830 & 0.803 & 0.960 & 0.970 & 0.992 & 0.988 & 0.946 & 0.989 \\
& VGG-11    & 0.932 & 0.925 & 0.939 & 0.956  & 0.940 & 0.950 & 0.949 & 0.961 & 0.995 & 0.985 & 0.950 & 0.989 \\
\cmidrule(lr){2-14} 
& {Average} & 0.891 & 0.888 & 0.902 & 0.884  & 0.902 & 0.892 & 0.951 & 0.965 & \textcolor{red}{0.992} & \textcolor{darkgreen}{0.986} & 0.950 & \textcolor{blue}{0.990} \\
\midrule
\multirow{5}{*}[3pt]{\rotatebox[origin=c]{90}{CUB-200}}  
& ResNet-50  & 0.861 & 0.850 & 0.851 & 0.780 & 0.804 & 0.911 & 0.458 & 0.967 & 0.989 & 0.975 & 0.925 & 0.952 \\
& ResNet-101 & 0.533 & 0.543 & 0.461 & 0.797 & 0.543 & 0.806 & 0.724 & 0.966 & 0.987 & 0.948 & 0.924 & 0.940 \\
& PMG        & 0.923 & 0.913 & 0.970 & 0.812 & 0.889 & 0.949 & 0.740 & 0.978 & 0.990 & 0.944 & 0.978 & 0.966 \\
\cmidrule(lr){2-14}
& {Average} & 0.772 & 0.769 & 0.761 & 0.796 & 0.745 & 0.889 & 0.641 & \textcolor{darkgreen}{0.970} & \textcolor{red}{0.989} & \textcolor{blue}{0.956} & 0.942 & {0.953} \\
\midrule
\multirow{7}{*}[3pt]{\rotatebox[origin=c]{90}{ImageNet-C}}  
& ViT               & 0.970 & 0.958 & 0.977 & 0.670 & 0.970 & 0.936 & 0.885 & 0.868 & 0.991 & 0.984  & 0.902 & 0.970\\
& DenseNet          & 0.963 & 0.957 & 0.977 & 0.976 & 0.964 & 0.981 & 0.855 & 0.970 & 0.995 & 0.990  & 0.908 & 0.990\\
& ConvNeXt          & 0.543 & 0.355 & 0.409 & 0.269 & 0.543 & 0.391 & 0.813 & 0.918 & 0.967 & 0.957  & 0.734 & 0.449\\
& CLIP-ViT-B        & 0.930 & 0.931 & 0.958 & 0.883 & 0.931 & 0.861 & 0.915 & 0.971 & 0.989 & 0.991  & 0.884 & 0.986\\
& CLIP-ConvNeXt     & 0.964 & 0.957 & 0.976 & 0.758 & 0.964 & 0.900 & 0.898 & 0.927 & 0.964 & 0.973  & 0.882 & 0.981\\
\cmidrule(lr){2-14}
& {Average}         & 0.876 & 0.831 & {0.859} & 0.711 & 0.874 & 0.814 & 0.873 & \textcolor{blue}{0.931} & \textcolor{red}{0.981} & \textcolor{darkgreen}{0.979}  & 0.862 & {0.875}\\
\midrule
\multirow{7}{*}[3pt]{\rotatebox[origin=c]{90}{ImageNet-3D}}  
& ViT              & 0.983 & 0.956 & 0.991 & 0.081 & 0.982 & 0.893 & 0.903 & 0.821 & 0.975 & 0.966 & 0.795 & 0.966 \\
& DenseNet         & 0.972 & 0.932 & 0.989 & 0.807 & 0.972 & 0.950 & 0.707 & 0.881 & 0.977 & 0.971 & 0.757 & 0.963 \\
& ConvNeXt         & 0.969 & 0.939 & 0.982 & 0.794 & 0.969 & 0.936 & 0.772 & 0.791 & 0.976 & 0.961 & 0.589 & 0.962 \\
& CLIP-ViT-B       & 0.941 & 0.898 & 0.989 & 0.912 & 0.938 & 0.962 & 0.933 & 0.943 & 0.976 & 0.963 & 0.890 & 0.964 \\
& CLIP-ConvNeXt    & 0.914 & 0.853 & 0.973 & 0.857 & 0.903 & 0.932 & 0.940 & 0.941 & 0.970 & 0.962 & 0.910 & 0.938 \\
\cmidrule(lr){2-14}
& {Average}        & 0.956 & 0.916 & \textcolor{red}{0.985} & 0.690 & 0.953 & 0.934 & 0.903 & 0.875 & \textcolor{darkgreen}{0.975} & \textcolor{blue}{0.964} & 0.840 & 0.958 \\
\midrule
\multicolumn{2}{c|}{Average over all setups}  & 0.863 & 0.851 & 0.891 & 0.795 & 0.865 & 0.882 & 0.914 & 0.935 & \textcolor{red}{0.984} & \textcolor{darkgreen}{0.971} & 0.899 & \textcolor{blue}{0.964} \\
\bottomrule
\end{tabular}
\end{center}
\end{table*}

\section{Dataset-Centric View: Unsupervised Model Evaluation}
\label{sec:data}
\noindent \textbf{Task Definition.}
Due to distribution shift ($p_S \neq p_T$), the accuracy of a model on the in-distribution test set $\mathcal{D}_\mathrm{test}^{S}$ is generally a poor indicator of its performance on the target (out-of-distribution) distribution $p_T$. This work aims to assess the generalization ability of a source-trained model $f$ on the target distribution $p_T$ \emph{without access to any labels}.
Concretely, given a model $f$ trained on labeled data from the source distribution $p_S$, and an unlabeled test set $\mathcal{D}_\mathrm{u}^{T} = \{\vx_i^{t}\}_{i=1}^{n_t}$ with $n_t$ i.i.d.\ samples drawn from $p_T$, the goal is to design a quantity that correlates strongly with the true classification accuracy of $f$ on $\mathcal{D}_\mathrm{u}^{T}$. We operate in the \emph{closed-set setting}, where the source and target distributions share the same set of $k$ classes.
Unlike domain adaptation, which focuses on adapting the model to improve its performance on the target distribution, our objective is purely \emph{evaluative}: we aim to predict the model's performance on various unlabeled test sets, without modifying the model or requiring access to labels.

\noindent \textbf{Evaluation Procedure.} Given a trained classifier, we test it on all the test sets under each setup. For each test set, we calculate the ground-truth accuracy and the estimated OOD quantity. Then, we evaluate the correlation strength between the estimated OOD quantity and accuracy. We also show scatter plots and mark real-world datasets for comparison. 

\noindent  \textbf{Evaluation Metrics.} To measure the quality of estimations, we use Pearson Correlation coefficient ($r$)~\cite{benesty2009pearson} and Spearman's Rank Correlation coefficient ($\rho$)~\cite{kendall1948rank} to quantify the linearity and monotonicity, respectively.
They range from $[-1, 1]$. A value closer to $1$ (or $-1$) indicates a strong positive (or negative) correlation, and $0$ implies no correlation \cite{benesty2009pearson}.
To precisely show the correlation, we use prob axis scaling that maps the range of both accuracy and estimated OOD quantity from $[0,1]$ to $[-\infty, +\infty]$, following~\cite{taori2020measuring, miller2021accuracy}. 
We also report the coefficient of determination ($R^2$) \cite{nagelkerke1991note} of the linear fit between estimated OOD quantity and accuracy following \cite{yu2022predicting}. The coefficient $R^2$ ranges from $0$ to $1$. An $R^2$ of $1$ indicates that the regression predictions perfectly fit OOD accuracy.

\begin{figure*}[t]
    \centering
\includegraphics[width=\linewidth]{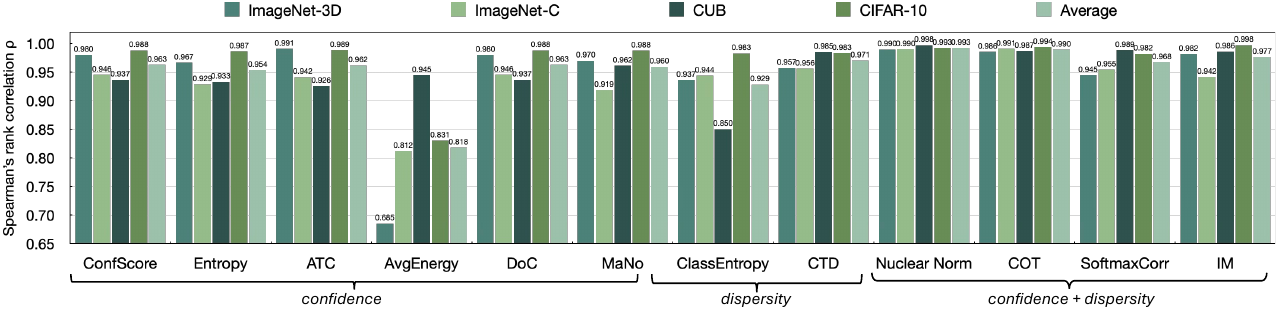}
    \caption{
\textbf{Average Spearman's Rank Correlation coefficient $\rho$ of each metric across ImageNet-C, ImageNet-3D, CIFAR-10, and WILDS setups in dataset-centric evaluation.}  
Each bar shows the correlation ($\rho$) between a metric and model accuracy across multiple test sets.
Metrics are grouped into three categories: 
(i) Confidence-based (\textit{e.g.}, ATC, DoC), which may perform well on clean data but degrade under distribution shift;
(ii) Dispersity-based (\textit{e.g.}, ClassEntropy, CTD), which capture prediction variation across classes and are more robust across datasets;
(iii) Hybrid metrics (\textit{e.g.}, NuclearNorm, COT), which combine confidence and dispersity and consistently achieve the strongest alignment with true accuracy rankings.
\textbf{Note:} CTD and COT are negatively correlated with accuracy by design, so $-\rho$ is shown for comparability.
}
    \label{fig:rho_evaluation}
\end{figure*}

\subsection{Experimental Setups} \label{sec:exp_setting}

\paragraph{ImageNet-1K} \underline{(i) Model.} We use $5$ representative neural networks provided by \cite{rw2019timm}. We include vision transformer ViT-Base-P16 (ViT) \cite{dosovitskiy2020image} and two convolution neural networks, DenseNet-121 (DenseNet) and ConvNeXt-Base \cite{liu2022convnet}. 
They are either trained or fine-tuned on ImageNet training set~\cite{deng2009imagenet}.
To assess the generalization of all methods, we also include two zero-shot vision-language models: CLIP-ViT-B/32 and CLIP-ConvNeXt-Base~\cite{radford2021learning}.

\underline{(ii) Synthetic Corruption Shift.} We use ImageNet-C benchmark \cite{hendrycks2019robustness} to study the synthetic distribution shift. ImageNet-C is controllable in terms of both type and intensity of corruption. It contains $95$ datasets that are generated by applying  $19$ types of corruptions (\textit{e.g.}, blur and contrast) to the ImageNet validation set. Each type has five intensity levels.
\underline{(iii) Synthetic 3D Shift.} We use the 3D Common Corruptions (ImageNet-3D) benchmark \cite{kar20223d} to study realistic distribution shifts. Unlike ImageNet-C \cite{hendrycks2019robustness}, which applies 2D corruptions uniformly, ImageNet-3D leverages 3D scene information to simulate more plausible corruptions based on depth, geometry, and viewpoint. We evaluate six types of 3D corruptions: \textit{far focus}, \textit{near focus}, \textit{xy motion blur}, \textit{z motion blur}, \textit{flash}, and \textit{fog 3D}, each with five severity levels. These corruptions introduce depth-aware blurring, spatially varying illumination, and camera-induced occlusions, which better reflect real-world image degradations.
\underline{(iiv) Real-world Shift.} We consider four natural shifts, including 1) dataset reproduction shift in ImageNet-V2-A/B/C~\cite{recht2019imagenet}, 2) sketch shift in  ImageNet-S(ketch)~\cite{wang2019learning}, 3) style shift in ImageNet-R(endition)~\cite{hendrycks2021many}, and 4) bias-controlled dataset shift in ObjectNet~\cite{barbu2019objectnet}. 
Note that, ImageNet-R and ObjectNet only share common $113$ and $200$ classes with ImageNet, respectively. Following \cite{hendrycks2021many}, we sub-select the model logits for the common classes with the ImageNet validation set.

\paragraph{CIFAR-10}
\underline{(i) Model.} We use ResNet-20 \cite{he2016deep}, RepVGG-A0 \cite{ding2021repvgg}, and VGG-11 \cite{simonyan2014very}. They are trained on the CIFAR-10 training set.
\underline{(ii) Synthetic Shift.} Similar to ImageNet-C, we use CIFAR-10-C \cite{hendrycks2019robustness} to study the synthetic shift.
It contains $19$ types of corruption and each type has $5$ intensity levels.
\underline{(iii) Real-world Shift.} We include three test sets: 1) CIFAR-10.1 with reproduction shift  \cite{recht2018cifar}, 2) CIFAR-10.2 with reproduction shift~\cite{recht2018cifar}, and 3) CINIC-10 that is sampled from a different database ImageNet.

\paragraph{CUB-200} We also consider fine-grained categorization with large intra-class variations and small inter-class variations~\cite{wei2021fine}.
We build up a setup based on the CUB-200-2011 dataset~\cite{wah2011caltech} that contains 200 birds categories.
\underline{(i) Model.} We use $3$ classifiers: ResNet-50,  ResNet-101, and PMG \cite{du2020fine}. They are pretrained on ImageNet and finetuned on the CUB-200-2011 training set. We use the publicly available codes provided by \cite{du2020fine}.
\underline{(ii) Synthetic Shift.} Following the protocol in ImageNet-C, we create CUB-200-C by applying $19$ types of corruptions with $5$ intensity levels to CUB-200-2011 test set.
\underline{(iii) Real-world Shift.} We use CUB-200-P(aintings) with style shift \cite{wang2020progressive}. It contains bird paintings with various renditions (\textit{e.g.,} watercolors, oil paintings, pencil drawings, stamps, and cartoons) collected from the web.

\subsection{Observations and Analysis}
Based on the results in Table~\ref{tab:corr_rho} (the coefficient of determination ($R^2$)) and Figure~\ref{fig:rho_evaluation} (Spearman’s $\rho$), we draw the following observations.

\paragraph{Confidence-based metrics show promising results on certain architectures, but their generality remains limited} As shown in Table~\ref{tab:corr_rho}, metrics such as ATC and DoC achieve high $R^2$ scores on models like VGG-11 (\textit{e.g.}, 0.939 for ATC on CIFAR-10) and ViT (0.991 for ACT on ImageNet-3D). However, their performance can degrade for other architectures; for instance, DoC drops to 0.543 on ResNet-101 in CUB-200 and 0.543 on ConvNeXt in ImageNet-C. This variance is further reflected in Figure~\ref{fig:rho_evaluation}. These observations suggest that the confidence signal can be informative under certain configurations, but its effectiveness is not universally stable across models and tasks.

\paragraph{Dispersity-based metrics offer more consistent and architecture-agnostic performance} CTD and ClassEntropy achieve high alignment across all model types. For example, CTD reaches an average $R^2$ of 0.970 on CUB-200 and 0.965 on CIFAR-10, while ClassEntropy also ranks highly across diverse architectures. Their usefulness is also evident in Figure~\ref{fig:rho_evaluation}, where both CTD and ClassEntropy maintain reasonably good rank correlation across datasets. 

\paragraph{Hybrid metrics that combine confidence and dispersity consistently outperform single-property approaches}
NuclearNorm, COT, and IM consistently achieve strong agreement with ground-truth accuracy across diverse architectures, particularly when confidence-only metrics are less effective.SoftmaxCorr, while slightly more variable, still maintains robust performance and often outperforms confidence-only or dispersity-only metrics. As also shown in Fig.~\ref{fig:rho_evaluation}, hybrid methods achieve consistently high Spearman correlations, confirming the advantage of jointly modeling confidence and dispersity for unsupervised model evaluation.

\begin{figure*}[!ht]
    \begin{center}
    \includegraphics[width=1\linewidth]{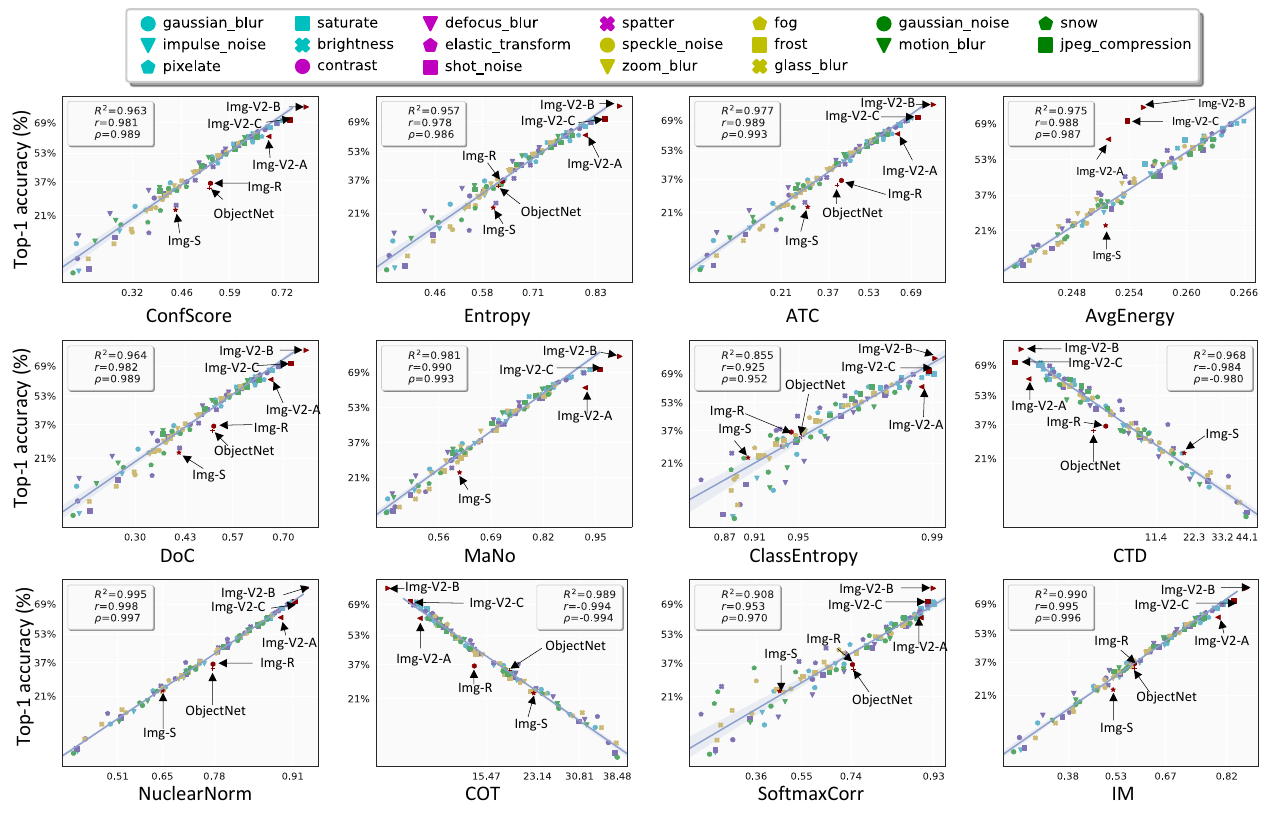}
    \caption{\textbf{Correlation study under the ImageNet setup.}
    We plot the actual accuracy of \textbf{\textit{DenseNet}} against predictions from \textbf{12} methods. Each shape in a subfigure denotes a test set, and the solid lines represent linear fits on synthetic ImageNet-C subsets. 
    We mark six real-world datasets with arrows and summarize the 19 ImageNet-C corruption types at the top using distinct shape–color pairs. 
    While metrics (\textit{e.g.}, ConfScore and ClassEntropy) exhibit noisy trends, COT and especially NuclearNorm show strong, consistent alignment with accuracy, with NuclearNorm yielding the closest fit to the regression line.}
    \label{fig:imagenet}
    \end{center}
\end{figure*}

\begin{figure*}[!ht]
    \begin{center}
    \includegraphics[width=1\linewidth]{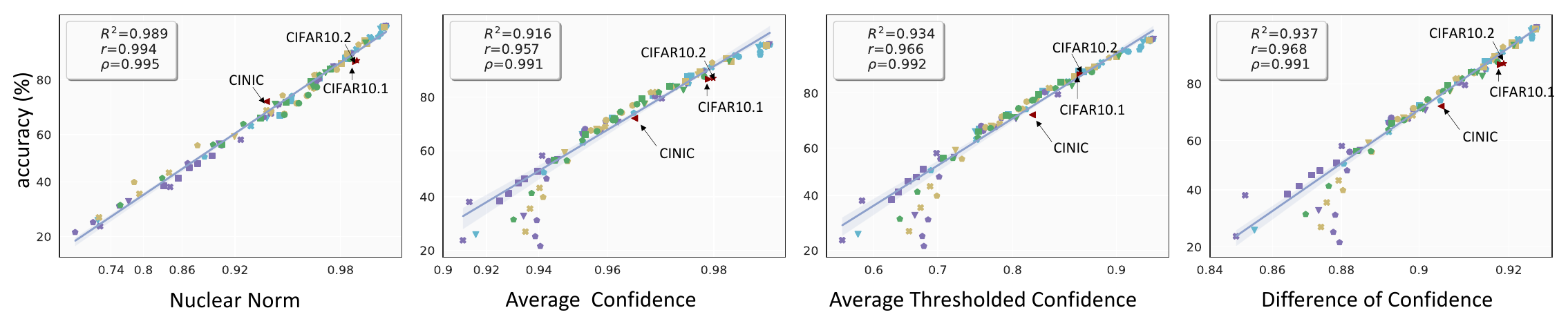}
    \caption{\textbf{Correlation study under the CIFAR-10 setup.} We plot the actual accuracy of \textbf{\textit{ResNet-20}} and the estimated OOD quantity. We show the results of nuclear norm, ConfScore, ClassEntropy, and COT.
   The lines are calculated by the linear regression fit on CIFAR-C. 
    We mark the $3$ real-world test sets in each sub-figure.
    }\label{fig:cifar}
    \end{center}
\end{figure*}

\begin{figure*}[t]
    \begin{center}
    \includegraphics[width=1\linewidth]{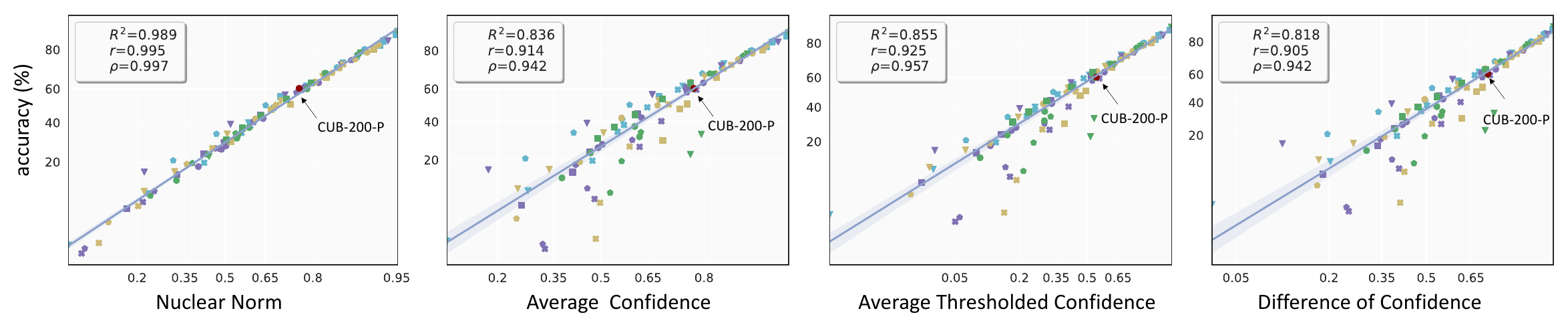}
    \caption{\textbf{Correlation study under the CUB-200 setup.}
    We plot the actual accuracy of \textbf{\textit{ResNet-50}} and the estimated OOD quantity. We show results of nuclear norm, ConfScore, ClassEntropy, and COT
    The straight lines are calculated by the linear regression fit on CUB-200-C. 
    We mark the real-world test set CUB-P in each sub-figure.
    }\label{fig:cub}
    \end{center}
\end{figure*}

\begin{figure*}[t]
    \begin{center}
    \includegraphics[width=1\linewidth]{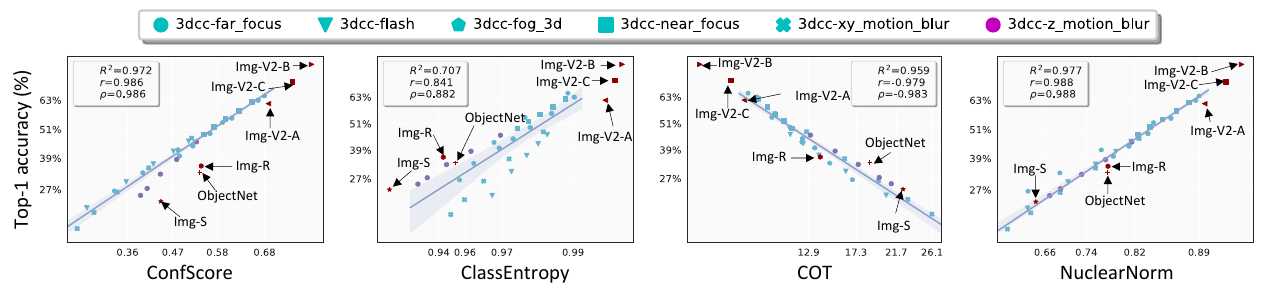}
    \caption{\textbf{Correlation study under the ImageNet-3D setup.}
    We visualize the actual accuracy of \textbf{\textit{DenseNet}} against the estimated OOD generalization signals from four metrics: NuclearNorm, ConfScore, ClassEntropy, and COT. 
    We mark six real-world datasets with arrows and summarize the six ImageNet-3D corruption types at the top using distinct shape–color pairs. 
    Each subplot includes a linear regression line fitted on the ImageNet-3D test sets, and real-world test sets are explicitly marked.
    }\label{fig:imagenet-3d}
    \end{center}
\end{figure*}

\paragraph{Nuclear norm can estimate the accuracy of real-world datasets}
We visualize the predictions of nuclear norms on real-world datasets in ImageNet (Fig.~\ref{fig:imagenet}), CIFAR-10 (Fig.~\ref{fig:cifar}), CUB-200 (Fig.~\ref{fig:cub}), and ImageNet-3D setups (Fig.~\ref{fig:imagenet-3d}). In all cases, nuclear norm aligns closely with ground-truth accuracy, placing real-world test sets near the regression line fitted on synthetic shifts. For example, under the ImageNet setup, it accurately predicts performance on ImageNet-V2-A/B/C, while other methods like ATC and DoC deviate on ImageNet-S and ObjectNet. Similar patterns hold for CIFAR-10 (Fig.~\ref{fig:cifar}), CUB-200 (Fig.~\ref{fig:cub}), and ImageNet-3D setups~(Fig.~\ref{fig:imagenet-3d}), where other metrics tend to underestimate accuracy on harder test sets. Compared to confidence-based and dispersity-based baselines, the nuclear norm provides more stable and reliable estimates across diverse distribution shifts.

\begin{figure*}[t]
    \begin{center}
    \includegraphics[width=1.0\linewidth]{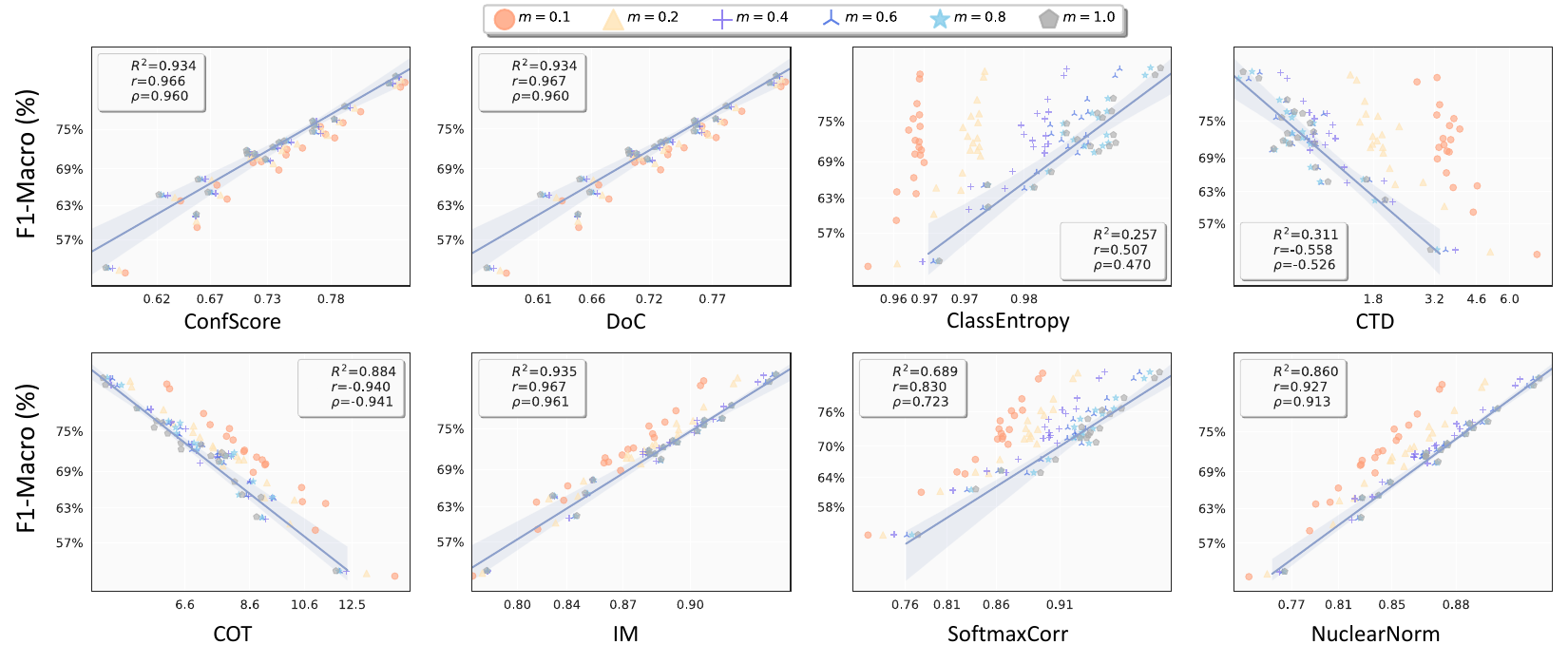}
    \caption{\textbf{Robustness of various methods to class imbalance.} 
   Using \textbf{\textit{ViT}} under the ImageNet setup, we evaluate the robustness of eight methods across different imbalance ratios $m$ in long-tailed test sets. 
    A smaller $m$ indicates higher imbalance severity. Linear regression lines are fit using the balanced test set ($m=1$).
    We find that {ConfScore} and {DoC} consistently exhibit strong correlation with actual performance across all imbalance levels. 
    In contrast, dispersity-based metrics ({ClassEntropy} and {CTD}) show weaker and unstable predictive alignment.
    Hybrid metrics ({MI}, {NuclearNorm}, {COT}, and {SoftmaxCorr}) show reduced reliability under strong imbalance ($m<0.4$), but remain robust when the imbalance is mild ($m \geq 0.4$).
    }\label{fig:imb}
    \end{center}
\end{figure*}

\paragraph{Discussion on class imbalance}
We construct long-tailed versions of ImageNet-C using exponential decay~\cite{cao2019learning}, with the imbalance ratio $m$ denoting the proportion between the least and most frequent class. We evaluate five imbalance levels: $\{0.1, 0.2, 0.4, 0.6, 0.8\}$. As shown in Fig.~\ref{fig:imb}, we observe three distinct behavioral groups.

First, {ConfScore} and {DoC} remain consistently aligned with ground-truth accuracy across all imbalance levels and demonstrate strong robustness. Second, the two dispersity-only metrics, {ClassEntropy} and {CTD}, perform poorly across all settings, exhibiting weak and noisy correlations. Third, hybrid metrics including {MI}, {NuclearNorm}, {COT}, and {SoftmaxCorr} show reduced reliability under severe imbalance ($m < 0.4$), but remain effective when the imbalance is mild ($m \geq 0.4$).

This resilience under mild imbalance arises from their design. {COT} aligns predicted class distributions with a uniform prior via Wasserstein distance, preserving stability when inter-class relations are retained. {SoftmaxCorr} captures class co-occurrence via second-order correlation, offering robustness under moderate skew. {NuclearNorm} evaluates the global structure of the prediction matrix without relying on class priors, encouraging confident and well-distributed predictions. {IM} combines marginal and average entropy to reflect both confidence and class spread.

Moreover, prediction dispersity remains a valuable signal even under strong imbalance, provided that the label distribution is known or can be estimated. Rather than assuming a uniform prior, adapting metrics like IM to account for target-aware priors could improve their robustness. This opens a promising direction for future work that incorporates label shift estimation~\cite{garg2023rlsbench,lipton2018detecting,tian2020posterior} and prior-aware modeling~\cite{chen2021mandoline,sun2022prior}.

\begin{table*}[t]
    \centering
    \setlength{\tabcolsep}{1.5pt}
    \small
    \caption{
    \textbf{Comparison of 12 unsupervised metrics across ImageNet, CIFAR-10, and WILDS in  model-centric ranking task.} 
We report Spearman’s rank correlation ($\rho$) between each metric and ground-truth accuracy, grouped into confidence-based, dispersity-based, and hybrid categories. Among confidence-based metrics, ATC and DoC perform well on ImageNet and CIFAR-10 but show reduced performance on WILDS due to domain complexity. CTD performs relatively well among dispersity-based metrics, while ClassEntropy struggles on class-imbalanced datasets such as iWildCam. Hybrid metrics such as NuclearNorm, COT, and SoftmaxCorr consistently rank among the top-performing methods. IM is less effective on WILDS due to its sensitivity to label imbalance.
The best, second-best, and third-best metrics in each row are highlighted in \textcolor{red}{red}, \textcolor{darkgreen}{green}, and \textcolor{blue}{blue}, respectively.
\textbf{Note:} CTD and COT are negatively correlated with accuracy (higher values indicate lower accuracy). To ensure consistency, we report $-\rho$ so that higher values always indicate better agreement.
    }
        \begin{tabular}{c c|cccccc|cc|cccc}
    \toprule
    \multirow{2}{*}{\textbf{Setup}}& \multirow{2}{*}{\textbf{Dataset} }
    & \multicolumn{6}{c|}{\textbf{Confidence}} 
    & \multicolumn{2}{c|}{\textbf{Dispersity}} 
    & \multicolumn{4}{c}{\textbf{Confidence + Dispersity}} \\
    \cmidrule(lr){3-8} \cmidrule(lr){9-10} \cmidrule(lr){11-14}
    & & ConfScore & Entropy & ATC & AvgEnergy & DoC & MaNo 
      & ClassEntropy & CTD 
      & NuclearNorm & COT & SoftmaxCorr & IM \\
    \midrule
    \multirow{8}{*}[3pt]{\rotatebox[origin=c]{90}{ImageNet}}  
    & ImageNet-V2      & 0.709 & 0.400 & 0.994 & -0.230 & 0.988 & 0.566 & 0.483 & 0.506 & 0.939 & 0.828 & 0.921 & 0.507 \\
    & ImageNet-S       & 0.862 & 0.741 & 0.981 & 0.400  & 0.955 & 0.834 & 0.836 & 0.978 & 0.975 & 0.951 & 0.935 & 0.930 \\
    & ObjectNet        & 0.883 & 0.821 & 0.962 & 0.372  & 0.926 & 0.896 & 0.752 & 0.917 & 0.952 & 0.917 & 0.963 & 0.924 \\
    & ImageNet-Blur    & 0.816 & 0.717 & 0.937 & 0.308  & 0.916 & 0.831 & 0.783 & 0.966 & 0.961 & 0.951 & 0.961 & 0.916   \\
    & ImageNet-A       & 0.754 & 0.609 & 0.828 & 0.377  & 0.880 & 0.781 & -0.554& -0.211 & 0.839 & 0.757 & 0.964 & 0.579 \\
    & ImageNet-R       & 0.828 & 0.699 & 0.950 & 0.510  & 0.953 & 0.898 & 0.610 & 0.740 & 0.942 & 0.872 & 0.951 & 0.919  \\
    \cmidrule(lr){2-14}
    & Average           & 0.809 & 0.665 & \textcolor{darkgreen}{0.942} & 0.290 & \textcolor{blue}{0.935} & 0.801 & 0.485 & 0.647 & \textcolor{blue}{0.935} & 0.879 & \textcolor{red}{0.949}  & 0.796   \\
    \midrule
    \multirow{5}{*}[2pt]{\rotatebox[origin=c]{90}{CIFAR-10}}
    & CIFAR-10.1        & 0.833 & 0.791 & 0.992 & 0.324 & 0.969 & 0.827 & 0.765 & 0.847 & 0.879 & 0.867 & 0.898 & 0.825  \\
    & CIFAR-10.2        & 0.833 & 0.791 & 0.992 & 0.324 & 0.968 & 0.825 & 0.918 & 0.953 & 0.885 & 0.872 & 0.894 & 0.856  \\
    & CINIC             & 0.651 & 0.609 & 0.949 & 0.481 & 0.849 & 0.654 & 0.851 & 0.869 & 0.727 & 0.705 & 0.821 & 0.740  \\
    & CIFAR-10-Noise    & 0.049 & 0.023 & 0.220 & 0.634 & 0.228 & 0.186 & 0.810 & 0.959 & 0.939 & 0.955 & 0.931 & 0.839  \\
    \cmidrule(lr){2-14}
    & Average           & 0.592 & 0.553 & 0.788 & 0.442 & 0.753 & 0.623 & 0.836 & \textcolor{red}{0.907} & \textcolor{blue}{0.858} & 0.850 & \textcolor{darkgreen}{0.886} & 0.815  \\
    \midrule
    \multirow{5}{*}[2pt]{\rotatebox[origin=c]{90}{WILDS}}
    & Camelyon17-OOD    & 0.192  & 0.167 & 0.111 & 0.323 & 0.046 & 0.175 & 0.581 & 0.572 & 0.772 & 0.682 & 0.630 & 0.618 \\
    & DomainNet-OOD     & 0.598  & 0.554 & 0.706 & 0.473 & 0.684 & 0.623 & 0.632 & 0.885 & 0.919 & 0.896 & 0.855 & 0.834 \\
    & iWildscam-OOD     & 0.912  & 0.847 & 0.835 & 0.374 & 0.911 & 0.931 & -0.415& 0.827 & 0.876 & 0.864 & 0.619 & -0.190 \\
    \cmidrule(lr){2-14}
    & Average           & 0.567  & 0.523 & 0.551 & 0.398 & 0.547 & 0.576 & 0.266 & \textcolor{blue}{0.761} & \textcolor{red}{0.856} & \textcolor{darkgreen}{0.814} & 0.701 & 0.421 \\
    \midrule
    \multicolumn{2}{c|}{Average over all setups}  & 0.656 & 0.580 & 0.760 & 0.416 & 0.746 & 0.667 & 0.529 & 0.781 & \textcolor{red}{0.883} & \textcolor{darkgreen}{0.848} & \textcolor{blue}{0.845} & 0.677 \\
    \bottomrule
    \end{tabular}
    \label{tab:all-metrics-ranking}
\end{table*}

\section{Model-Centric View: Unsupervised Model Ranking}
\label{sec:model}
\noindent \textbf{Task Definition.}
We study the problem of ranking pretrained classifiers on an unlabeled OOD test set. 
Suppose we are given a set of $M$ models $\{\phi_1, \dots, \phi_M\}$, each trained independently on $\mathcal{D}^S$. For each model $\phi_m$, we compute its prediction matrix $\bm{P}_m \in \mathbb{R}^{N \times K}$ by applying the model to each test input $x_i \in \mathcal{D}^T$. 
While the true accuracy of each model on $\mathcal{D}^T$ is unknown due to the lack of ground-truth labels, our objective is to estimate the rankings of all models. Specifically, we aim to construct a score function that maps each $\bm{P}_m$ to a scalar score $S_m$, such that the scores $\{S_m\}_{m=1}^M$ preserve the relative ranking of models' actual performance. This setting defines the task of \emph{unsupervised model selection}, where no access to test labels is assumed.

\noindent \textbf{Evaluation metrics.} We use Spearman's Rank Correlation coefficient $\rho$ \cite{kendall1948rank} to measure monotonicity between calculated scores and model accuracy. We also compute the weighted variant of Kendall’s rank correlation $\tau_w$, which is shown to be a useful measure when selecting the best-ranked item of interest \cite{you2021logme}. Both range from $[-1, 1]$. A value closer to $-1$ or $1$ indicates a strong negative or positive correlation, respectively, and $0$ means no correlation. Similar to \cite{miller2021accuracy} and \cite{baek2022agreement}, we apply the same probit scale to both accuracy and SoftmaxCorr in our experiment for a better linear fit.

\subsection{Experimental Setup}
\paragraph{ImageNet setup} 
We collect 180 models publicly accessible from TIMM \cite{rw2019timm}. They are trained or fine-tuned on ImageNet \cite{deng2009imagenet} and have various architectures, training strategies, and training paradigms. In addition to models that are trained on the ID training dataset, we also consider $90$ zero-shot vision-language models, including CLIP~\cite{radford2021learning}, SigLIT~\cite{zhai2023sigmoid}, BLIP~\cite{li2022blip}, BLIP-2~\cite{li2023blip2} and Flava \cite{singh2022flava}. We use the default prompt set for corresponding models. If the default prompt sets are not provided, ``\texttt{A picture of \{class\}.}'' is deployed.
We use five OOD datasets for the correlation study: (1) ImageNet-V2 \cite{recht2019imagenet}; (2) ObjectNet \cite{barbu2019objectnet}; (3) ImageNet-S(ketch) \cite{wang2019learning}; (4) ImageNet-Blur severity $5$ \cite{hendrycks2019benchmarking}; (5) ImageNet-R(endition) \cite{hendrycks2021many}; ImageNet-R and ObjectNet contain $200$ and $113$ ImageNet classes, respectively. 
We use Top-1 accuracy as a metric for classification.

\paragraph{CIFAR-10 setup}
We collect $65$ networks trained with the scheme provided by \cite{kl2017cifar} on CIFAR-10 training set \cite{krizhevsky2009learning}. These models have different model architectures. CIFAR-10-Val(idation) is the ID test set. For OOD datasets, we use (1) CIFAR-10.2 \cite{recht2018cifar10.1}, which is the reproduction of CIFAR-10 by extracting $2,000$ images from TinyImage. 
(2) CINIC \cite{darlow2018cinic}, which is an extended alternative for CIFAR-10. It is collected by combining CIFAR-10 with images selected and down-sampled from ImageNet. 
(3) CIFAR-10-Noise with severity~$5$~\cite{hendrycks2019benchmarking}, which is created by artificially corrupting CIFAR-10-Val with a Gaussian noise function, and it has $10,000$ images in each CIFAR-10 class.
We use accuracy as the metric of model generalization.

\paragraph{WILDS setup} We consider a classification tasks of this setup: Camelyon17 \cite{bandi2018detection}.
It is a binary classification dataset where the objective is to classify whether a slide contains a tumor issue. We use $45$ models varying in architectures and random seeds. 
ID and OOD datasets are the default ID validation set and OOD test set, respectively. 
For DomainNet \cite{peng2019moment}, we use publicly available model checkpoints, which are trained using the schema provided in \cite{sagawa2021extending}.
{iWildCam} is a $182$-way animal classification dataset. We collect $66$ models whose variation results from different network architectures and learning rates. Model performance is measured by macro-$F1$ score for both tasks.
For each task, we follow the same training scheme provided by \cite{koh2021wilds} to train or fine-tune models.

\begin{figure*}[!ht]
    \centering
\includegraphics[width=\linewidth]{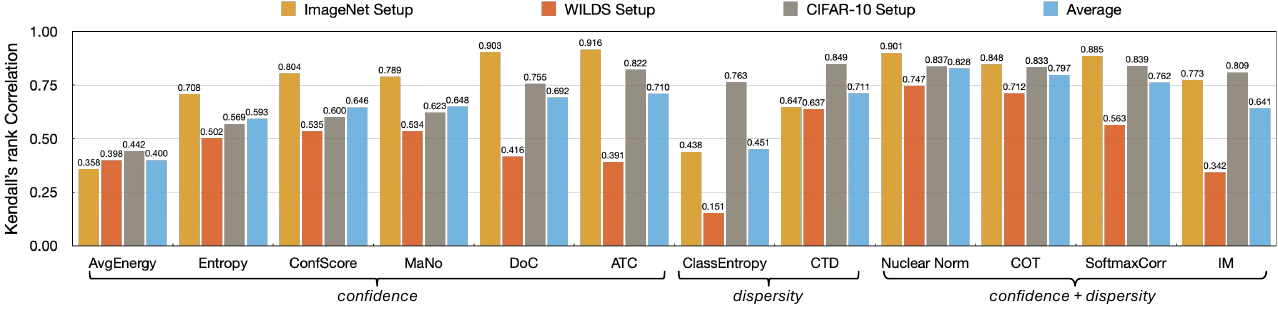}
  \caption{
\textbf{Average Kendall’s rank correlation $\tau_w$ of each metric across ImageNet, CIFAR-10, and WILDS setups in model-centric evaluation.}
Each bar shows how well a metric ranks multiple models on a fixed test set; higher values indicate stronger agreement with ground-truth accuracy.
Metrics are categorized into three groups:
(i) Confidence-based metrics (\textit{e.g.}, ConfScore, MaNo, ATC) perform well on clean datasets like ImageNet and CIFAR-10 but often degrade in WILDS due to domain-specific shifts.
(ii) Dispersity-based metrics (\textit{e.g.}, ClassEntropy, CTD) are more robust to distribution shifts but may fail under strong class imbalance.
(iii) Hybrid metrics (\textit{e.g.}, NuclearNorm, COT, SoftmaxCorr) consistently achieve high correlation across all setups by jointly modeling confidence and dispersity, while IM shows instability in WILDS due to its dependence on balanced class distributions.
\textbf{Note:} CTD and COT are distance-based and inherently negatively correlated with accuracy; we report $-\tau_w$ to ensure higher values consistently indicate stronger agreement.
}
    \label{fig:tau_ranking}
\end{figure*}

\begin{figure*}[!ht]
    \centering
\includegraphics[width=1\linewidth]{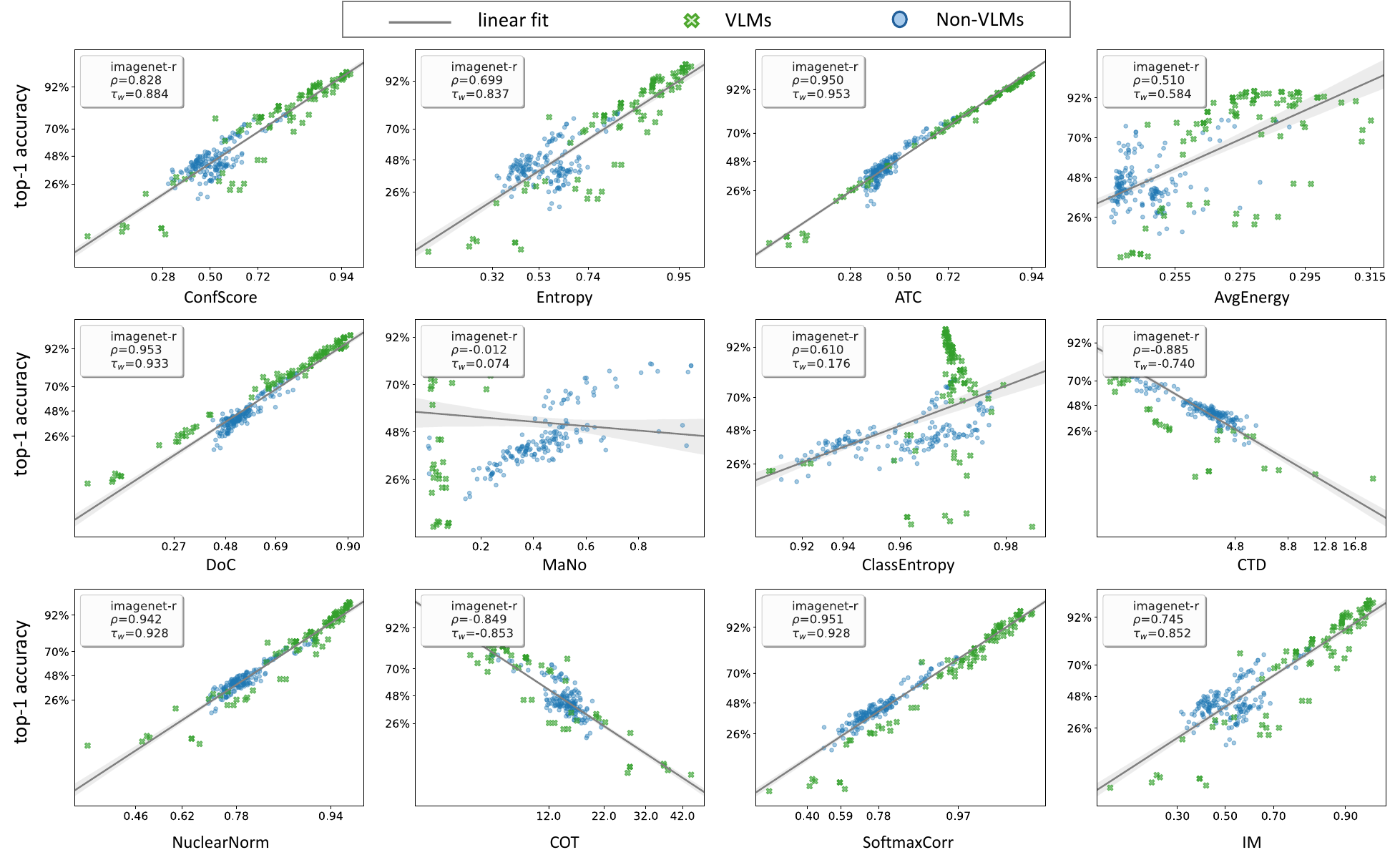}
\caption{
\textbf{Scatter plots of model-centric ranking performance on ImageNet-R.}
Each subplot shows the correlation between a metric (x-axis) and top-1 accuracy (y-axis) across a range of models. Vision-language models (VLMs) and non-VLMs are shown separately, and a linear fit (black line) is provided for reference. The Spearman’s rank correlation ($\rho$) and Kendall’s $\tau_w$ between the metric and accuracy are shown in each plot. Metrics capturing both confidence and dispersity, such as {NuclearNorm}, {SoftmaxCorr}, and {COT}, show strong and linear alignment across both VLM and non-VLM groups.
}
\label{fig:ranking-img-r}
\end{figure*}

\begin{figure*}[!ht]
    \centering
\includegraphics[width=1\linewidth]{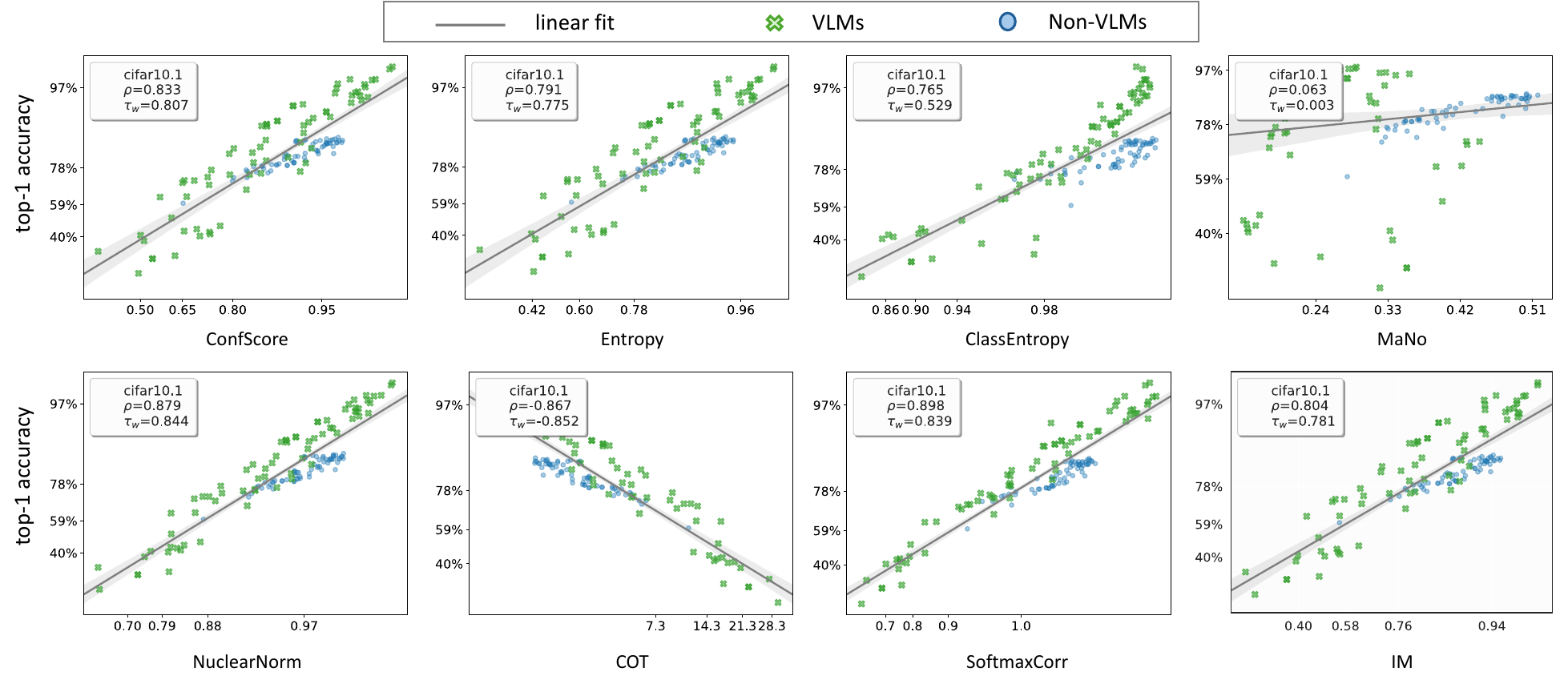}
\caption{
\textbf{Scatter plots of model-centric ranking performance on CIFAR-10.1.}
Each subplot shows the correlation between a metric (x-axis) and top-1 accuracy (y-axis) across a range of models. Vision-language models and non-VLMs are shown separately, with a linear fit (black) and corresponding Spearman’s $\rho$ and Kendall’s $\tau_w$. The patterns largely align with prior findings, suggesting that NuclearNorm, SoftmaxCorr, and COT consistently align with accuracy trends.
}
\label{fig:ranking-cifar}
\end{figure*}

\begin{figure*}[!ht]
    \centering
\includegraphics[width=1\linewidth]{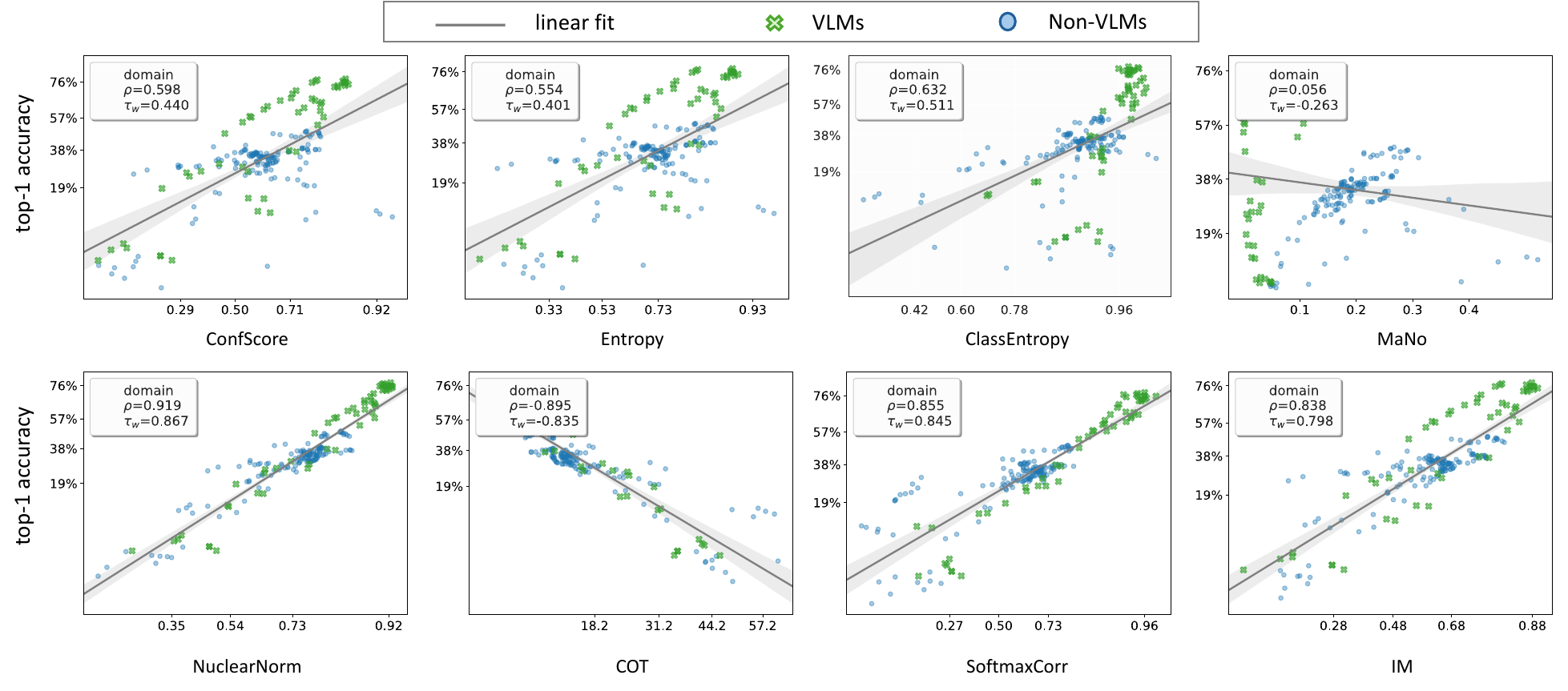}
\caption{
\textbf{Scatter plots of model-centric ranking performance on DomainNet-OOD.}
Each plot shows the correlation between a metric (x-axis) and top-1 accuracy (y-axis) across models, separated by vision-language and non-VLM groups. A linear fit (black) and rank correlations (Spearman’s $\rho$, Kendall’s $\tau_w$) are shown. The patterns largely align with prior findings, suggesting that NuclearNorm, SoftmaxCorr, and COT consistently align with accuracy trends.
}
\label{fig:ranking-domain}
\end{figure*}

\subsection{Observations and Analysis}
Based on the results in Table~\ref{tab:corr_rho} (Spearman’s $\rho$) and Figure~\ref{fig:rho_evaluation} (Spearman’s $\rho$), we draw three major observations.

\paragraph{First, confidence-based metrics demonstrate varying levels of effectiveness, with threshold-dependent methods showing sensitivity to validation–test domain shift} 
ATC and DoC rely on thresholds calibrated from a validation set, performing well when the validation and test distributions are aligned—achieving Kendall’s $\tau_w$ (Figure~\ref{fig:tau_ranking}) of 0.916 and 0.903 on ImageNet, and 0.822 and 0.755 on CIFAR-10. However, their performance drops significantly on WILDS (0.391 and 0.416), where domain shifts cause misaligned thresholds. Spearman’s $\rho$ (Table~\ref{tab:all-metrics-ranking}) shows a consistent trend: ATC and DoC fall from 0.942 and 0.935 on ImageNet to 0.551 and 0.547 on WILDS. This decline reveals the challenge of transferring threshold-dependent metrics across domains. In contrast, the other confidence-based metrics that do not require a validation set exhibit more stable, though generally weaker, correlation. Their average $\tau_w$ for three sets of controls ranges from 0.400 to 0.648, reflecting their limited capacity to account for the global class prediction structure.

\paragraph{Second, dispersity-based metrics provide useful distribution-level information, but their performance depends on class balance assumptions} 
These metrics assess the spread or concentration of predicted class probabilities and can offer strong correlation in well-balanced datasets. For example, on CIFAR-10, CTD and ClassEntropy achieve Kendall’s $\tau_w$ of 0.849 and 0.763, and Spearman’s $\rho$ of 0.907 and 0.836, respectively. However, both ClassEntropy and CTD explicitly assume uniform class distributions, which leads to poor generalization in imbalanced settings. On WILDS, where class imbalance is more severe, ClassEntropy and CTD achieve low Kendall’s $\tau_w$ (0.151 and 0.637) and degraded $\rho$ values. These results suggest that while dispersity captures valuable global cues, its effectiveness is limited when its underlying assumptions are violated.

\paragraph{Third, hybrid metrics that combine prediction confidence and dispersity consistently outperform single-aspect metrics} 
NuclearNorm, COT, and SoftmaxCorr all rank among the top-performing methods across ImageNet, CIFAR-10, and WILDS. NuclearNorm achieves the highest average Kendall’s $\tau_w$ across all three setups (0.901, 0.837, 0.747) and leads Spearman’s $\rho$ with an average of 0.883.
In addition, we present scatter plots for Imagenet-R (Fig.~\ref{fig:ranking-img-r}), CIFAR-10.1 (Fig.~\ref{fig:ranking-cifar}), and DomainNet-OOD (Fig.~\ref{fig:ranking-domain}), comparing unsupervised metric scores and top-1 accuracy across different models. Hybrid metrics like NuclearNorm, COT, and SoftmaxCorr exhibit strong correlation with accuracy. In contrast, metrics like MaNo and ClassEntropy show weak or inconsistent alignment, particularly for non-VLMs.

\paragraph{Resilience of Hybrid Metrics to Class Imbalance}
While IM combines entropy and marginal entropy directly and is sensitive to class imbalance, COT, NuclearNorm, and SoftmaxCorr adopt more resilient designs.
COT and SoftmaxCorr use fixed reference priors (e.g., uniform or identity), but incorporate global class-level structure—via transport consistency and class correlation respectively—which helps offset imbalance effects.
NuclearNorm avoids assuming any class prior and instead encourages confident yet diverse predictions across classes. These properties enable the three hybrid metrics to maintain strong performance on class-imbalanced datasets such as ImageNet-A and iWildscam

\section{Conclusion and Discussion}
This work presents a unified framework for unsupervised model assessment, covering two practical tasks: dataset-centric evaluation, which estimates the accuracy of a fixed model on multiple unlabeled test sets, and model-centric ranking, which identifies the most suitable model from a pool of candidates for a given unlabeled dataset. These scenarios frequently arise in real-world applications where labeled test data is not available.
While most prior efforts rely on prediction confidence as the primary signal of generalization, we revisit the role of prediction dispersity, which reflects how predictions are distributed across output classes. We demonstrate that confidence and dispersity each capture important and complementary aspects of model behavior. 
To this end, we systematically benchmark a range of unsupervised metrics, including confidence-based, dispersity-based, and combined approaches, across diverse datasets, architectures, and distribution shifts. Our results show that metrics that integrate both prediction confidence and dispersity offer more stable and reliable generalization estimates. In particular, the nuclear norm of the prediction matrix consistently performs well across both evaluation and ranking tasks. 
We also examine its robustness under class imbalance and find it remains effective under moderate shifts, though sensitivity may arise in more extreme cases.
Overall, our findings support the value of jointly modeling confidence and dispersity when evaluating model performance without labels. This contributes to a deeper understanding of generalization in unlabeled environments and offers useful guidance for model assessment in practical deployment scenarios.

\noindent\textbf{Limitation and Future Work.} The current framework, while providing robust unsupervised assessment for classification, primarily focuses on tasks with {categorical outputs} and relies on the explicit structure of the {softmax prediction matrix}. This design choice limits its direct applicability to broader machine learning domains where output spaces differ. Specifically, {the methodology does not immediately generalize to regression tasks}~\cite{thiagarajan2024pager} or {complex structured prediction settings} (\textit{e.g.}, object detection~\cite{yang2024bounding,yu2024towards,hu2024towards} and graph data~\cite{lu2024temporal}), where outputs are continuous or spatially correlated. A key practical constraint is the assumption of access to full model outputs ({softmax probabilities}), which is {often unavailable in resource-constrained or privacy-sensitive black-box deployment scenarios}. Furthermore, {robustness challenges under severe label shift} warrant future investigation, as our analysis revealed that dispersity-based and hybrid metrics can exhibit {reduced reliability under strong class imbalance}. Addressing these limitations presents a rich agenda for future work, notably by extending the evaluation to support non-categorical outputs and developing reliable methods for black-box assessment.

\section*{Acknowledgement}
This research was funded by the National Road Safety Program under the Australian Government’s Department of Infrastructure, Transport, Regional Development, Communications and the Arts (Grant No. NRSAGP-TI1-A48). The authors gratefully acknowledge this funding support.

\bibliographystyle{IEEEtran}
\bibliography{reference}

\end{document}

%% file: math_command.tex
\usepackage{amsmath,amsfonts,bm}

\def\eqref#1{equation~\ref{#1}}

\def\1{\bm{1}}

\def\vx{{\bm{x}}}

\def\mP{{\bm{P}}}

\DeclareMathAlphabet{\mathsfit}{\encodingdefault}{\sfdefault}{m}{sl}
\SetMathAlphabet{\mathsfit}{bold}{\encodingdefault}{\sfdefault}{bx}{n}